\documentclass[runningheads]{llncs}

\usepackage{eccv}

\usepackage{eccvabbrv}

\usepackage{graphicx}
\usepackage{booktabs}
\usepackage{multirow}

\usepackage{graphicx}
\usepackage{subcaption}

\usepackage[table]{xcolor}
\usepackage{xfp}

\usepackage{siunitx}
\newcommand{\fixedone}[1]{\num[round-precision=1,minimum-decimal-digits=1]{#1}}
\newcommand{\fixedtwo}[1]{\num[round-precision=2,minimum-decimal-digits=2]{#1}}

\newcommand{\nochange}[1]{\textcolor{gray}{\tiny\,$\sim$#1}}
\newcommand{\upgood}[1]{\textcolor{red}{\tiny\,\,$\uparrow$#1}}
\newcommand{\downgood}[1]{\textcolor{green!55!black}{\tiny\,\,$\downarrow$#1}}

\newcommand{\autodelta}[1]{%
  \begingroup
  \edef\delta{\fpeval{#1}}%
  \ifnum\inteval{\fpeval{abs(\delta) < 0.05}}=1
    \nochange{\fixedone{\delta}}%
  \else
    \ifnum\inteval{\fpeval{\delta > 0}}=1
      \upgood{\fixedone{\delta}}%
    \else
      \downgood{\fixedone{\fpeval{abs(\delta)}}}%
    \fi
  \fi
  \endgroup
}

\newcommand{\upgoodHB}[1]{\textcolor{red}{\tiny\,\,$\uparrow$#1}}
\newcommand{\downgoodHB}[1]{\textcolor{green!55!black}{\tiny\,\,$\downarrow$#1}}

\newcommand{\upbadLB}[1]{\textcolor{green!55!black}{\tiny\,\,$\uparrow$#1}}
\newcommand{\downgoodLB}[1]{\textcolor{red}{\tiny\,\,$\downarrow$#1}}

\newcommand{\autodeltaoneLB}[1]{%
  \begingroup
  \edef\delta{\fpeval{#1}}%
  \ifnum\inteval{\fpeval{abs(\delta) < 0.05}}=1
    \nochange{\fixedone{\delta}}%
  \else
    \ifnum\inteval{\fpeval{\delta > 0}}=1
      \upbadLB{\fixedone{\delta}}%
    \else
      \downgoodLB{\fixedone{\fpeval{abs(\delta)}}}%
    \fi
  \fi
  \endgroup
}

\newcommand{\autodeltatwoHB}[1]{%
  \begingroup
  \edef\delta{\fpeval{#1}}%
  \ifnum\inteval{\fpeval{abs(\delta) < 0.005}}=1
    \nochange{\fixedtwo{\delta}}%
  \else
    \ifnum\inteval{\fpeval{\delta > 0}}=1
      \upgoodHB{\fixedtwo{\delta}}%
    \else
      \downgoodHB{\fixedtwo{\fpeval{abs(\delta)}}}%
    \fi
  \fi
  \endgroup
}

\usepackage[accsupp]{axessibility}  % Improves PDF readability for those with disabilities.

\usepackage{hyperref}

\usepackage{orcidlink}

\begin{document}

\title{Decoding by Perturbation: Mitigating MLLM Hallucinations via Dynamic Textual Perturbation} 

\titlerunning{DeP}

\author{Sihang Jia\inst{1}\thanks{Equal contribution} \and
Shuliang Liu\inst{1}\protect\footnotemark[1] \and
Songbo Yang\inst{1} \and
Yibo Yan\inst{1} \and
Xin Zou\inst{1} \and
Xuming Hu\inst{1}\thanks{corresponding author}
}

\authorrunning{S.~Jia, S.~Liu, et al.}

\institute{The Hong Kong University of Science and Technology (Guangzhou)\\
\email{sjia188@connect.hkust-gz.edu.cn, xuminghu@hkust-gz.edu.cn}}

\maketitle

\begin{abstract}
  Multimodal Large Language Models frequently suffer from inference hallucinations, partially stemming from language priors dominating visual evidence. Existing training-free mitigation methods either perturb the visual representation and deviate from the natural image distribution, or enforce intrusive manipulations that compromise the model’s inherent generative fluency. We introduce a novel perspective that multimodal hallucination manifests as the hypersensitivity of visual grounding to textual phrasing during the decoding phase. Building on this insight, we propose \textbf{Decoding by Perturbation (DeP)}, a training-free framework mitigating prior-induced hallucinations via controlled textual interventions. DeP employs a dynamic probe applying multi-level textual perturbations to elicit latent language priors. Leveraging attention variance, it enhances stable evidence regions while suppressing suspicious noise in the feature space. Furthermore, it constructs an interpretable prior drift direction using logits statistics to counteract probability biases from textual co-occurrences. Extensive experiments confirm DeP effectively reduces hallucinations and achieves superior performance across multiple benchmarks.
\end{abstract}

\section{Introduction}
\label{sec:intro}

\begin{figure}
    \centering
    \includegraphics[width=0.9\linewidth]{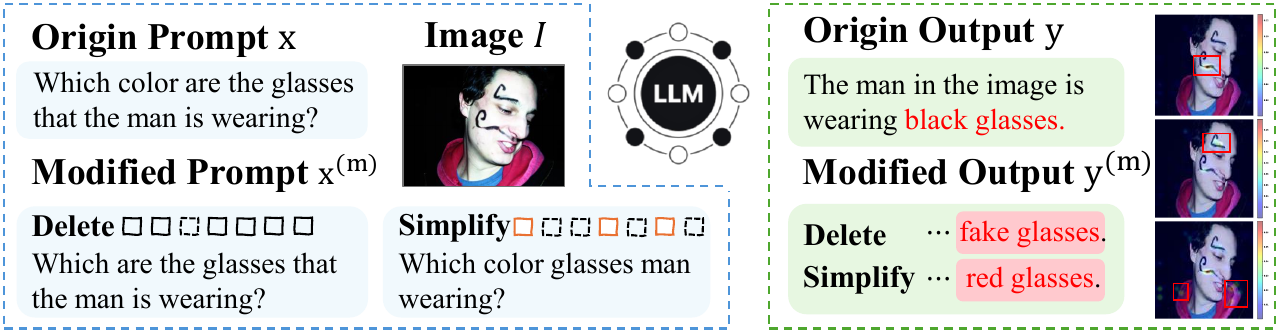}
    \caption{Illustration of multimodal hallucination driven by text hypersensitivity. Although the modified prompts preserve the core semantics of the original query, minor variations in their surface structures lead to drastically different hallucinated outputs (e.g., ``fake'' or ``red''). As revealed by the attention maps on the right, these textual perturbations cause a severe drift in visual grounding. Notably, the generation of ``red'' indicates that the model has established a spurious correlation, erroneously shifting its visual attention (bottom map) to the red hoodie rather than grounding it on the glasses.}
    \label{fig:intro_pic}
\end{figure}

Recent advancements in Multimodal Large Language Models (MLLMs) have demonstrated significant progress across tasks such as visual question answering, image captioning, and visual dialogue. Despite their robust visual comprehension capabilities, these models frequently suffer from multimodal hallucination during inference \cite{yang2025inex}. This phenomenon manifests as generated responses containing inaccurate descriptions of the input image, including references to non-existent objects or erroneous spatial relationships. Prior research attributes the origins of multimodal hallucination to multifaceted factors, ranging from perceptual limitations in visual encoders and insufficient modal alignment to inherent biases within language models \cite{an2025mitigating,fang2025grounding,guan2024hallusionbench}
% jo2026attention,li2025mitigating,liu2024phd,min2024mitigating,pan2026ground,park2025halloc,tang2025seeing,wu2025antidote,yang2025nullu}.
A critical underlying mechanism involves the imbalance between visual and linguistic weights during decoding. Given the strong autoregressive inertia of Large Language Models (LLMs), the generation process often succumbs to statistical co-occurrences within the input text prompts \cite{jo2026attention,wu2025antidote,liu2025survey}. This linguistic dominance suppresses the constraints of visual evidence when processing key tokens, causing the generation trajectory to be mainly pulled by the text prompt and resulting in descriptions detached from the image content.

To mitigate this issue, existing strategies primarily fall into training-based methods \cite{sun2024aligning,zhao2023beyond,xiao2025detecting,zhao2025mitigating,zhang2025mm,ding2025pami,hesystematic,liu2026distilling} and training-free inference-time interventions \cite{li2025mitigating,li2025vidhalluc,liu2026vision,min2024mitigating,pan2026ground,shi2026one,tang2025seeing,wu2025generate,xiao2026macd,yang2025inex,yang2025nullu,zhang2025self,zhang2026hallucination,zhao2025aligning,zhao2025cross,zhu2025ibd,zheng2026visual}. Due to the prohibitive computational costs of the former, the plug-and-play nature of the latter has garnered significant attention. Current inference-time interventions generally follow two paths. First, decoding strategy of visual intervention creates visual uncertainty by constructing image degradation samples \cite{an2025mitigating,chen2025decoupling,chen2025ict,lyu2024alleviating,xiao2026macd}, and uses the difference information between the original view and the degraded view to suppress hallucinations \cite{chen2025mitigating,cho2025you,li2025mitigating,yin2025clearsight,zhang2025self,zhao2025cross}. However, this approach fundamentally alters the raw characteristics of the input data, causing the model to deviate from its authentic inference mode under natural image distributions and hindering the precise localization of hallucination triggers \cite{an2025mitigating,chen2025decoupling,lyu2024alleviating,wu2025generate,zhao2025cross}. Second, attention interventions attempt to directly rectify the imbalance between vision and language within the decoding layers. Nevertheless, such intrusive parameter adjustments often lack a robust decoupling mechanism, potentially compromising the fluent generation capabilities of the LLM while suppressing hallucinations \cite{chen2025decoupling,jo2026attention,zhao2025aligning}. Consequently, we center our research on a fundamental attribution decoupling problem: how can we precisely isolate false generation tendencies induced by textual conditions without compromising the integrity of visual evidence? \cite{duan2025truthprint}

We address this challenge by introducing a novel perspective where multimodal hallucination during inference is essentially characterized by the hypersensitivity of visual attention to textual phrasing, as illustrated in Fig.~\ref{fig:intro_pic}. Provided the core semantics of the prompt remain consistent, the model's attention to visual evidence should exhibit stability despite variations in the surface structure of the text. Conversely, hallucinations often stem from the decoder's over-reliance on language priors, establishing spurious statistical mappings between specific syntactic structures and image noise. Therefore, the key to identifying hallucinations lies in probing whether the model's visual grounding maintains feature alignment robustness under linguistic perturbations.

Guided by this insight, and distinguishing our approach from traditional methods that disrupt image structure, we propose \textbf{Decoding by Perturbation (DeP)}, a training-free decoding calibration framework tailored to mitigate language prior-driven hallucinations via multi-level text perturbations. DeP leverages the statistical characteristic of textual sensitivity to elicit implicit linguistic prior dependencies while keeping the image representation invariant. The framework operates through a closed loop of detection and correction comprising three synergistic modules \cite{duan2025truthprint}. First, during decoding, we dynamically generate a set of semantics-preserving perturbed prompts based on the uncertainty of the current step to trigger potential language priors. Second, utilizing attention statistics across these perturbations, we explicitly decouple image regions into evidence areas that remain stable across textual conditions and suspicious areas that drift with textual induction \cite{liu2026vision}. We enhance the visual features of stable regions within the hidden state space while suppressing noise interference from suspicious regions, forcing the model decision to revert to visual facts. Finally, addressing linguistic biases that cannot be mapped to specific regions, we model the distributional difference between original and perturbed logits as a prior drift direction. We directly neutralize this drift in the output space to achieve interpretable, hallucination-free generation.

Our contributions are:
\begin{itemize}
    \item We model hallucination as the hypersensitivity of visual grounding to textual conditions and \textbf{for the first time} introduce \textbf{multi-text perturbation} as an observable probe to quantify this instability.
    \item We propose the training-free \textbf{DeP} framework, which achieves precise visual region decoupling and targeted linguistic bias elimination by statistically analyzing attention variance and logit drift directions under multiple perturbations.
    \item We conducted extensive experiments on multiple benchmark datasets and models, and achieves state-of-the-art performance on a range of evaluation metrics, verifying the effectiveness of our method in mitigating multimodal hallucinations.
\end{itemize}

\section{Related Work}

\paragraph{Multimodal Hallucination Mitigation.}
MLLMs are prone to generating content inconsistent with visual evidence \cite{liu2023visual, bai2024hallucination, fu2024mme, huo2025pmark, zhang2025catmark, zhang2025cohemark}. Existing mitigation strategies are broadly bifurcated into training-based alignment \cite{liu2024improved, chen2024sharegpt4v, liu2025vla} and training-free inference-time interventions \cite{leng2024mitigating, huang2024opera, yin2024woodpecker}. While training-based methods enhance factuality through instruction tuning or reinforcement learning, they incur significant computational overhead. In contrast, training-free interventions during decoding offer more flexible and efficient alternatives for real-time hallucination rectification.

\paragraph{Visual Degradation-based Interventions.}
This category suppresses language priors by introducing visual uncertainty to identify model over-reliance on text. VCD \cite{leng2024mitigating} and Med-VCD \cite{mahdavi2026med} utilize noise-augmented images for contrastive decoding, while CICD  and Scaffolding \cite{dang2025benchmark} introduce structure-disrupted or low-resolution views for calibration. Recent works such as Divter \cite{saxena2024mitigating}, P-VCD \cite{peng2025omnisync}, and R-VCD  further refine this by employing progressive visual corruption or multi-view consistency checks to isolate prior-induced tokens. Despite their utility, as highlighted in \cite{huang2024opera}, these methods rely on perturbing input images, which may fundamentally alter the raw visual distribution and hinder the precise localization of hallucination root causes.

\paragraph{Intrusive Interventions on Internal Representations.}
Another research line manipulates internal activations or attention mechanisms to rectify vision-language imbalances. OPERA \cite{huang2024opera} and ClearSight \cite{yin2025clearsight} introduce over-trust penalties and retrospection strategies within the attention flow. More recent intrusive methods like VTI \cite{xu2025rethinking}, Nullu \cite{yang2025nullu}, and ASD \cite{su2025activation} directly steer hidden states or modify latent space distributions during inference. Advanced techniques such as Skip-Attention , Layer-Wise Intervention \cite{chen2025attention}, and Steering-LLM \cite{sivakumar2025model} attempt to prune or edit internal weights to enhance visual grounding. However, such parameter-level adjustments often lack robust decoupling, potentially compromising the model's inherent generative fluency \cite{wang2025shift, fu2025lingoloop}. Our DeP framework differs by introducing a non-intrusive textual perturbation probe, preserving internal structural integrity.

\section{Methodology}

\begin{figure}
    \centering
    \includegraphics[width=0.9\linewidth]{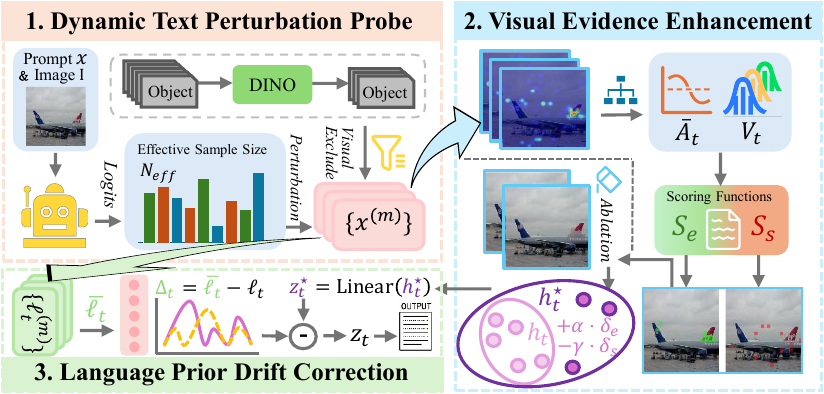}
    \caption{Overview of the DeP framework. DeP mitigates language prior-driven hallucinations during inference based on text perturbations. First, DeP utilizes attention consistency statistics across perturbations to decouple visual evidence and suspicious regions for hidden state calibration. Then, it estimates the prior drift direction from perturbed logits, applying it as a penalty to yield the final prediction.}
    \label{fig:main_pic}
\end{figure}

We introduce \textbf{Decoding by Perturbation (DeP)}, a training-free framework operating exclusively during inference to mitigate hallucinations driven by language priors in multimodal generation, illustrated in Fig.~\ref{fig:main_pic}. This method establishes an explicit axis for textual conditional intervention by applying controlled perturbation distributions to the text prompt $x$. This process deliberately relaxes visual constraints and amplifies the dominance of language priors. By observing the response of the logit distribution and cross-modal attention to these conditional variations at decoding step $t$, we quantify and subsequently correct this prior-driven tendency. The framework comprises three core modules: \textbf{Dynamic Text Perturbation Probe}, \textbf{Visual Evidence Decoupling and Enhancement}, and \textbf{Language Prior Drift Correction}.

\subsection{Dynamic Text Perturbation Probe}

At each decoding step $t$, relying solely on the original prompt $x$ makes it difficult to discern whether the generation stems from visual facts or linguistic inertia. To disentangle these factors, we employ \textbf{multiple textual perturbations} as probes. Unlike single perturbations, which may introduce accidental semantic drift, multiple perturbations allow us to transform sensitivity to textual conditions into estimable statistics. Specifically, given the original prompt $x$, we construct a perturbation set containing $M$ variations:
\begin{equation}
x^{(m)} \sim q(\tilde{x}\mid x),\quad m=1,\dots,M
\end{equation}
Conditioned on the fixed image $I$ and the preceding generated sequence $y_{<t}$, We obtain the logits and cross-modal attention tensors corresponding to the perturbed inputs:
\begin{equation}
\begin{aligned} \ell_t^{(m)} &= \log P_\theta(\cdot \mid I, x^{(m)}, y_{<t}) \\ A_t^{(m)} &= \text{Attention}(\cdot \mid I, x^{(m)}, y_{<t}) \end{aligned}
\end{equation}
To minimize uncontrolled fluctuations, we design the perturbation distribution $q(\cdot\mid x)$ to be highly controllable, deliberately steering it towards directions that elicit latent language priors. Specifically, it aims to satisfy \textbf{two criteria}: \textbf{1)} ensuring that prior-driven candidate tokens and visual attention exhibit heightened sensitivity under perturbation; and \textbf{2)} preventing significant semantic drift that could compromise the assessment of prior sensitivity.

\subsubsection{Multi-level Semantic Perturbation Strategy}

To activate potential language priors at various levels, we abstract text perturbation as a unified conditional lexical replacement framework. Given the original prompt sequence $x=[w_{1},...,w_{L}]$, we define a unified perturbation mapping function $\mathcal{T}_{k}$ as follows:
\begin{equation}
\mathcal{T}_{k}(x) = \text{Replace}(x; w \rightarrow w'), \quad \text{s.t.} \quad w \in \Omega_k, w' \sim q_k(\cdot|w)
\end{equation}
Here, $w$ denotes the target token selected based on strategy $\Omega_k$, and $w'$ represents the replacement word sampled from a specific distribution $q_k$. Based on this framework, we design three specific replacement strategies along a progressive axis of \textbf{semantics preservation versus perturbation intensity}:
\textbf{1)} Visual Attribute Weakening ($\mathcal{T}_{1}$), \textbf{2)} High-Frequency Prior Replacement ($\mathcal{T}_{2}$), \textbf{3)} Co-occurrence Adversarial Intervention ($\mathcal{T}_{3}$).
(Detailed design specifications are provided in the supplementary materials.)

% Visual Attribute Weakening ($\mathcal{T}_{1}$): This strategy primarily targets components with a weak visual basis in the prompt, such as adjectives, for lightweight perturbation. By reducing fine-grained visual constraints, we induce the model to rely on intrinsic language priors.

% High-Frequency Prior Replacement ($\mathcal{T}_{2}$): This strategy restricts perturbation targets to key entities, replacing them with high-frequency generic objects. The objective is to explicitly introduce a generic visual concept prior. 
% % When the model lacks sufficient evidence from the real image, language priors tend to supply answers based on high-frequency concepts. This strategy exacerbates such tendencies, exposing prior-driven candidate tokens.

% Co-occurrence Adversarial Intervention ($\mathcal{T}_{3}$): This strategy replaces key entities with adversarial concepts that possess high co-occurrence probabilities. The approach is more specifically designed to same-domain confusion, forcefully inducing the model to reveal its dependence on pure text statistical regularities.

\subsubsection{Distribution Shape-Driven Perturbation Mode Selection}

To maximize the prior sensitivity signal while maintaining semantic stability, we dynamically schedule perturbation strategies based on the shape of the logit distribution at the current decoding step. We first quantify the current prediction uncertainty of the model by defining the \textbf{Effective Sample Size} as the inverse square of the $L_2$ norm of the probability distribution $p_t = \text{softmax}(\ell_t)$:
\begin{equation}
N_{\mathrm{eff}}(p_t) = \|p_t\|_2^{-2} = \left(\sum_{v \in \mathcal{V}} p_{t,v}^2\right)^{-1}
\end{equation}
Intuitively, a larger $N_{\mathrm{eff}}$ suggests a dispersed distribution, reflecting model hesitation. Based on this metric, we construct a piecewise strategy selection function $\pi(x)$ to adaptively match the optimal probe intensity to the cognitive state of the model:
\begin{equation}
\pi(x)=\begin{cases}\mathcal{T}_{1}(x),&\text{if } N_{eff}(p_{t})\le\delta\quad(\text{High Certainty}),\\ \mathcal{T}_{3}(x),&\text{if } N_{eff}(p_{t})>\delta\quad(\text{High Uncertainty}).\end{cases}
\end{equation}
Finally, to ensure the rationality and feasibility of perturbations, we introduce two constraint mechanisms. 
\textbf{1) Strategy Fallback Mechanism:} If the adversarial mapping sampling for $\mathcal{T}_{3}(x)$ fails (i.e., $\mathcal{M}_{\text{adv}}(s) = \emptyset$), the strategy automatically falls back to $\mathcal{T}_{2}(x)$. This ensures a continuous injection of controlled external prior signals. 
\textbf{2) Visual Mutual Exclusion Constraint:} To prevent significant semantic drift caused by the replacement word $w'$ actually existing in the image, we employ a DINO \cite{zhang2022dino} detector to admit replacement words only if they satisfy the visual exclusion condition.
% \begin{equation}
% \mathbb{I}_{\text{valid}}(w') = \begin{cases} 1, & \text{if } \text{similarity}(\mathcal{D}(I), \text{emb}(w')) < 0.3, \\ 0, & \text{otherwise}. \end{cases}
% \end{equation}

\subsection{Visual Evidence Decoupling and Enhancement}

To transform textual condition sensitivity from a mere risk signal into actionable visual evidence that directly constrains generation, we utilize attention consistency statistics derived from multiple text perturbations. This allows us to explicitly decouple two types of regions in the image space: \textbf{evidence regions $\mathcal{R}_e$}, which remain stable across perturbations, and \textbf{suspicious regions $\mathcal{R}_s$}, which drift with textual guidance. We define the first moment (mean) and second central moment (variance) matrices of attention over the perturbation set as:
\begin{equation}
\bar{A}_t = \mathbb{E}_{m}[A_t^{(m)}], \qquad V_t = \mathbb{V}\text{ar}_{m}[A_t^{(m)}] = \mathbb{E}_{m}\left[ (A_t^{(m)} - \bar{A}_t)^{\circ 2} \right]
\end{equation}
Here, $\bar{A}_t$ characterizes consistent attention across textual conditions, while $V_t$ captures attention drift sensitive to language priors. We utilize $\bar{A}_t$ and $V_t$ for region separation, estimating the sensitivity contribution of evidence and suspicious regions to current generation, thereby calibrating the decoding process.

\subsubsection{Region Separation and Ablation}

Let $\mathcal{N}(\cdot)$ denote a normalization operator mapping matrices to $[0,1]$. We define:
\begin{equation}
\tilde{A}_t=\mathcal{N}(\bar{A}_t),\qquad \tilde{V}_t=\mathcal{N}(V_t),\qquad \tilde{\sigma}_t=\sqrt{\tilde{V}_t}.
\end{equation}
Building upon these normalized statistics, we introduce confidence bound corrections based on multi-perturbation sampling to construct scoring functions for both region types.

The \textbf{Evidence Score} is defined as:
\begin{equation}
S_e=\tilde{A}_t-\lambda_e(M,\delta_e)\cdot \tilde{\sigma}_t,
\end{equation}
where, $\lambda_e(M,\delta_e)$ is a conservative coefficient determined by the number of perturbation samples $M$ and confidence level $\delta_e$. This term is equivalent to a Lower Confidence Bound (LCB) estimate of stable attention intensity under limited sampling. It prioritizes visual evidence that receives sustained attention with low uncertainty across multiple perturbations, while suppressing unstable attention that drifts with specific lexical induction (indicated by increased $\tilde{\sigma}_t$).

The \textbf{Suspicious Score} is defined as:
\begin{equation}
S_s=\Big((1-\tilde{A}_t)-\lambda_a(M,\delta_a)\Big)\;\odot\;\Big(\tilde{\sigma}_t+\lambda_s(M,\delta_s)\Big).
\end{equation}
The first factor, $(1-\tilde{A}_t)-\lambda_a(M,\delta_a)$, represents the LCB for attention absence, favoring regions consistently lacking attention support. The second factor, $\tilde{\sigma}_t+\lambda_s(M,\delta_s)$, represents the Upper Confidence Bound (UCB) for instability, emphasizing regions likely to exhibit significant drift under text perturbation. This score identifies suspicious locations that display drastic drift while lacking stable attentional grounding.

Based on these scores, we obtain binary \textbf{masks $M_e$ and $M_s$} for the image. To test the model's dependence on visual details in these regions, we apply region-level information suppression. Given a region $R \in \{\mathcal{R}_e, \mathcal{R}_s\}$, the perturbed image $I_{\ominus R}$ is defined as a mixture of the original image and Gaussian convolution features:
\begin{equation}
I_{\ominus R} = (1 - M_R) \odot I + M_R \odot (G_\sigma * I),
\end{equation}
This operation applies Gaussian blur only to pixels covered by the mask, weakening identifiable details within the region while preserving global structure and context.

\subsubsection{Region Sensitivity Estimation and Feature Calibration}

At the same decoding step $t$, we perform forward passes on $I$, $I_{\ominus R_e}$, and $I_{\ominus R_s}$ to obtain hidden states. We define the region sensitivity vector as the difference between the original and blurred states:
\begin{equation}
\delta_e = h_t(I) - h_t(I_{\ominus R_e}), \qquad \delta_s = h_t(I) - h_t(I_{\ominus R_s})
\end{equation}
Here, $\delta_e$ reflects the model's reliance on visual information in evidence regions, while $\delta_s$ reflects reliance on suspicious regions. Finally, we perform contrastive fusion in the hidden state space to obtain the calibrated hidden state \cite{hu2022hiure}:
\begin{equation}
h_t^{\star} = h_t(I) + \alpha \cdot \delta_e - \gamma \cdot \delta_s
\end{equation}
Hyperparameters $\alpha$ and $\gamma$ control the intensity of evidence enhancement and suspicious suppression, respectively. The calibrated state $h_t^{\star}$ explicitly injects region-level signals into the current language decision, biasing subsequent sampling toward candidates supported by stable visual evidence and suppressing deviations caused by prior traction.

\subsection{Language Prior Drift Correction}

While we utilize attention consistency differences to map prior-dominated signals to image regions for visual alignment, certain prior biases manifest directly in token preferences within the language modality. Therefore, by leveraging logits statistics from multiple perturbations, we construct an interpretable prior shift direction and counteract it in the final logits.

At decoding step $t$, let the logits obtained from the forward pass under the original condition be $\ell_t \in \mathbb{R}^{|\mathcal{V}|}$. For the $M$ text perturbation prompts $\{x^{(m)}\}_{m=1}^{M}$, we obtain the corresponding logits:
\begin{equation}
\ell_t^{(m)}=\log p(\cdot \mid I, x^{(m)}, y_{<t})
\end{equation}
We calculate the expectation of the perturbed logits and the directional difference relative to the original logits:
\begin{equation}
\bar{\ell}_t=\frac{1}{M}\sum_{m=1}^{M}\ell_t^{(m)}, \qquad \Delta_t = \bar{\ell}_t - \ell_t
\end{equation}
Let the region-calibrated logits be $z_t^{\star} = \text{Linear}(h_t^{\star})$. We incorporate the directional difference as a penalty term in the final logits:
\begin{equation}
z_t = z_t^{\star} - \beta \cdot \Delta_t
\end{equation}
The hyperparameter $\beta \ge 0$ represents the directional guidance strength. Intuitively, $\Delta_t$ describes the consistency shift direction of the model's output distribution after perturbing text conditions, effectively representing the sensitivity direction toward language priors at the current step. This penalty term counteracts drifts caused by text priors in the output space, aligning the final distribution more consistently with stable decisions supported by visual evidence.

\section{Experiments}

\subsection{Experimental Setting}

\subsubsection{Benchmarks}
To rigorously evaluate hallucinations across different cognitive granularities, we adopt two standard benchmarks representing discriminative object detection and generative reasoning, respectively.

As a discriminative benchmark, \textbf{POPE} \cite{li2023evaluating} assesses the capability of models to verify the existence of objects in images through yes-or-no question answering. We conduct evaluations under all three settings: Random, Popular, and Adversarial, comprising 9,000 images and 27,000 questions in total. Following the standard protocol, we report accuracy, precision, recall, and F1 score to measure the consistency of object localization.

To evaluate hallucinations in more complex free-form generation scenarios, we utilize \textbf{MMHal-Bench} \cite{sun2024aligning}. Comprising 96 curated image-question pairs across 8 diverse categories (e.g., attribute and spatial reasoning), MMHal-Bench requires detailed sentence-level descriptions, making it highly sensitive to subtle attribute or relationship hallucinations. Following the official evaluation protocol, we employ GPT-4 as the judge to compare model responses with ground truth references. The judge scores the responses on a scale of 0 to 6 based on factual correctness and informativeness, and calculates the hallucination rate (a score below 3 indicates hallucinated generation).

\subsubsection{Implementation Details}
We apply DeP to two representative multimodal large language models: \textbf{LLaVA-1.5} \cite{liu2023llava} and \textbf{InstructBLIP} \cite{dai2023instructblip}. By default, we use greedy decoding to verify its generality. DeP operates exclusively during the inference stage and does not modify pre-trained weights. For the dynamic text perturbation probe, we set the number of perturbations to $M=3$ to balance estimation stability and inference latency. For the visual evidence enhancement module, the sensitivity control parameters are set to $\alpha=0.2$ and $\gamma=0.3$. In the language prior drift correction, the guidance strength is set to $\beta=2.0$.

\subsubsection{Baselines}
To comprehensively evaluate the effectiveness of DeP, we select five state-of-the-art training-free baseline methods for comparison. We categorize these methods into two main technical routes: (i) visual degradation-based intervention methods that suppress hallucinations by constructing visual uncertainty or contrastive views, including VCD \cite{leng2024mitigating} and CICD \cite{zhao2025cross}; (ii) internal representation-based intrusive intervention methods that attempt to directly modify attention weights or latent space distributions at the decoding layer, including ClearSight \cite{yin2025clearsight}, VTI \cite{liu2025reducing} and Nullu \cite{yang2025nullu}. For these baseline methods, we adopt the parameter settings reported in their respective papers.

\subsection{Main Results}

\begin{table}[t]
\centering
\caption{Performance evaluation on POPE (\%). Delta denotes the improvement compared to Origin. Best results are highlighted in \textbf{bold}.}
\resizebox{\textwidth}{!}{
\begin{tabular}{llcccccccc}
\toprule
\multirow{2}{*}{Setting} & \multirow{2}{*}{Method} &
\multicolumn{4}{c}{\textbf{LLaVA-1.5}} &
\multicolumn{4}{c}{\textbf{InstructBLIP}} \\
\cmidrule(lr){3-6}\cmidrule(lr){7-10}
& & Acc$\uparrow$ & Prec.$\uparrow$ & Rec.$\uparrow$ & F1$\uparrow$ &
Acc$\uparrow$ & Prec.$\uparrow$ & Rec.$\uparrow$ & F1$\uparrow$ \\
\midrule

\rowcolor{gray!15}
\multirow{7}{*}{Adversarial} 
& Origin
& 71.7\autodelta{0.0} & 65.7\autodelta{0.0} & 90.7\autodelta{0.0} & 76.2\autodelta{0.0}
& 68.5\autodelta{0.0} & 62.0\autodelta{0.0} & 81.5\autodelta{0.0} & 70.4\autodelta{0.0} \\
& VCD
& 70.4\autodelta{-1.3} & 64.5\autodelta{-1.2} & \textbf{92.5}\autodelta{+1.8} & 76.9\autodelta{+0.7}
& 70.7\autodelta{+2.2} & 64.9\autodelta{+2.9} & 84.1\autodelta{+2.6} & 73.2\autodelta{+2.8} \\
& CICD
& 72.0\autodelta{+0.3} & 66.1\autodelta{+0.4} & 92.0\autodelta{+1.3} & 77.0\autodelta{+0.8}
& 71.2\autodelta{+2.7} & 65.7\autodelta{+3.7} & \textbf{85.5}\autodelta{+4.0} & 74.3\autodelta{+3.9} \\
& ClearSight
& 74.3\autodelta{+2.6} & 68.8\autodelta{+3.1} & 91.1\autodelta{+0.4} & 78.3\autodelta{+2.1}
& 71.6\autodelta{+3.1} & 67.4\autodelta{+5.4} & 83.5\autodelta{+2.0} & 74.6\autodelta{+4.2} \\
& VTI
& 74.4\autodelta{+2.7} & 68.9\autodelta{+3.2} & 90.9\autodelta{+0.2} & 78.4\autodelta{+2.2}
& 73.8\autodelta{+5.3} & 68.7\autodelta{+6.7} & 85.4\autodelta{+3.9} & 76.1\autodelta{+5.7} \\
& Nullu
& 72.9\autodelta{+1.2} & 67.2\autodelta{+1.5} & 91.5\autodelta{+0.8} & 77.5\autodelta{+1.3}
& 72.3\autodelta{+3.8} & 68.9\autodelta{+6.9} & 82.9\autodelta{+1.4} & 75.3\autodelta{+4.9} \\
\rowcolor{gray!15}
& DeP(Ours)
& \textbf{77.4}\autodelta{+5.7} & \textbf{72.4}\autodelta{+6.7} & 90.1\autodelta{-0.6} & \textbf{80.3}\autodelta{+4.1}
& \textbf{76.4}\autodelta{+7.9} & \textbf{73.1}\autodelta{+11.1} & 85.2\autodelta{+3.7} & \textbf{78.7}\autodelta{+8.3} \\
\midrule

\rowcolor{gray!15}
\multirow{7}{*}{Popular}
& Origin
& 77.0\autodelta{0.0} & 71.1\autodelta{0.0} & 90.7\autodelta{0.0} & 79.7\autodelta{0.0}
& 74.5\autodelta{0.0} & 72.0\autodelta{0.0} & 83.9\autodelta{0.0} & 77.5\autodelta{0.0} \\
& VCD
& 77.8\autodelta{+0.8} & 72.5\autodelta{+1.4} & \textbf{92.5}\autodelta{+1.8} & 81.3\autodelta{+1.6}
& 75.0\autodelta{+0.5} & 71.6\autodelta{-0.4} & 85.2\autodelta{+1.3} & 77.8\autodelta{+0.3} \\
& CICD
& 79.3\autodelta{+2.3} & 74.2\autodelta{+3.1} & 92.1\autodelta{+1.4} & 82.2\autodelta{+2.5}
& 76.5\autodelta{+2.0} & 72.8\autodelta{+0.8} & 84.3\autodelta{+0.4} & 78.1\autodelta{+0.6} \\
& ClearSight
& 81.0\autodelta{+4.0} & 77.4\autodelta{+6.3} & 91.2\autodelta{+0.5} & 83.7\autodelta{+4.0}
& 78.7\autodelta{+4.2} & 76.0\autodelta{+4.0} & \textbf{86.6}\autodelta{+2.7} & 80.9\autodelta{+3.4} \\
& VTI
& 81.8\autodelta{+4.8} & 77.7\autodelta{+6.6} & 90.9\autodelta{+0.2} & 83.8\autodelta{+4.1}
& 79.2\autodelta{+4.7} & 76.2\autodelta{+4.2} & 85.4\autodelta{+1.5} & 80.5\autodelta{+3.0} \\
& Nullu
& 80.4\autodelta{+3.4} & 75.8\autodelta{+4.7} & 91.6\autodelta{+0.9} & 83.0\autodelta{+3.3}
& 77.6\autodelta{+3.1} & 74.2\autodelta{+2.2} & 83.2\autodelta{-0.7} & 78.4\autodelta{+0.9} \\
\rowcolor{gray!15}
& DeP(Ours)
& \textbf{82.8}\autodelta{+5.8} & \textbf{78.5}\autodelta{+7.4} & 92.0\autodelta{+1.3} & \textbf{84.7}\autodelta{+5.0}
& \textbf{81.2}\autodelta{+6.7} & \textbf{80.3}\autodelta{+8.3} & 85.3\autodelta{+1.4} & \textbf{82.7}\autodelta{+5.2} \\
\midrule

\rowcolor{gray!15}
\multirow{7}{*}{Random}
& Origin
& 84.1\autodelta{0.0} & 79.3\autodelta{0.0} & 90.8\autodelta{0.0} & 84.7\autodelta{0.0}
& 81.0\autodelta{0.0} & 79.7\autodelta{0.0} & 83.5\autodelta{0.0} & 81.6\autodelta{0.0} \\
& VCD
& 83.2\autodelta{-0.9} & 78.2\autodelta{-1.1} & \textbf{92.5}\autodelta{+1.7} & 84.8\autodelta{+0.1}
& 81.4\autodelta{+0.4} & 80.5\autodelta{+0.8} & 84.5\autodelta{+1.0} & 82.5\autodelta{+0.9} \\
& CICD
& 84.6\autodelta{+0.5} & 80.1\autodelta{+0.8} & 92.1\autodelta{+1.3} & 85.7\autodelta{+1.0}
& 83.1\autodelta{+2.1} & 81.6\autodelta{+1.9} & 84.1\autodelta{+0.6} & 82.8\autodelta{+1.2} \\
& ClearSight
& 86.8\autodelta{+2.7} & 82.9\autodelta{+3.6} & 91.2\autodelta{+0.4} & 86.9\autodelta{+2.2}
& 85.1\autodelta{+4.1} & 84.3\autodelta{+4.6} & 85.6\autodelta{+2.1} & 84.9\autodelta{+3.3} \\
& VTI
& 87.0\autodelta{+2.9} & 83.1\autodelta{+3.8} & 90.9\autodelta{+0.1} & 86.9\autodelta{+2.2}
& 85.4\autodelta{+4.4} & 84.6\autodelta{+4.9} & \textbf{86.4}\autodelta{+2.9} & 85.5\autodelta{+3.9} \\
& Nullu
& 85.7\autodelta{+1.6} & 81.5\autodelta{+2.2} & 91.6\autodelta{+0.8} & 86.3\autodelta{+1.6}
& 84.0\autodelta{+3.0} & 82.8\autodelta{+3.1} & 83.8\autodelta{+0.3} & 83.3\autodelta{+1.7} \\
\rowcolor{gray!15} 
& DeP(Ours)
& \textbf{87.2}\autodelta{+3.1} & \textbf{84.1}\autodelta{+4.8} & 92.2\autodelta{+1.4} & \textbf{88.0}\autodelta{+3.3}
& \textbf{85.9}\autodelta{+4.9} & \textbf{86.2}\autodelta{+6.5} & 85.4\autodelta{+1.9} & \textbf{85.8}\autodelta{+4.2} \\
\bottomrule
\end{tabular}
}
\label{tab:llava_instructblip_pope}
\end{table}

\subsubsection{Performance on POPE}
As shown in Tab.~\ref{tab:llava_instructblip_pope}, our training-free framework (DeP) significantly outperforms existing baseline methods across multiple metrics. DeP demonstrates exceptional performance improvements in the most challenging Popular and Adversarial settings. Under the Adversarial setting, DeP increases the F1 score of InstructBLIP from the baseline of \textbf{70.4} to \textbf{78.7}, achieving a substantial improvement of \textbf{8.3}. For LLaVA-1.5, DeP improves the F1 score from \textbf{79.7} to \textbf{84.7} under the Popular setting, and from \textbf{76.2} to \textbf{80.3} under the Adversarial setting. This proves the robustness of DeP when dealing with strong distractors.

Observing the performance of the original baseline models and other intervention methods reveals that the models often exhibit extremely high recall but relatively low precision. This occurs because the Popular and Adversarial test data in POPE deliberately trigger the inherent language priors of the models. Such priors cause the models to over-rely on text statistical patterns during decoding, leading to a biased tendency to blindly answer ``Yes''.
The DeP framework effectively suppresses this tendency. By substantially improving precision while maintaining a reasonable recall rate, DeP achieves a significant leap in the overall F1 score.

\begin{table}[t]
    \centering
    \caption{Performance evaluation on MMHal-Bench. \textbf{(a)} General results with delta improvements compared to Origin. \textbf{(b)} Detailed scores on the eight categories, where ``Overall'' indicates the averaged performance.}
    \label{tab_fig_combined_mmhal}

    \begin{subtable}[t]{0.63\textwidth}
        \centering
        \caption{General results on MMHal} % 空 caption 也会自动生成 (a)
        \vspace{0.4cm}
        \footnotesize
        \setlength{\tabcolsep}{2pt}
        \renewcommand{\arraystretch}{0.95}

        \begin{tabular}{llcc}
        \toprule
        \textbf{Model} & \textbf{Method} & Halluc. Rate$\downarrow$ & Score$\uparrow$ \\
        \midrule
        \multirow{7}{*}{\textbf{LLaVA-1.5}}
        & \cellcolor{gray!15}Origin & \cellcolor{gray!15}59.4\autodeltaoneLB{0.0} & \cellcolor{gray!15}2.54\autodeltatwoHB{0.00} \\
        & VCD & 61.5\autodeltaoneLB{+2.1} & 2.25\autodeltatwoHB{-0.29} \\
        & CICD & 58.3\autodeltaoneLB{-1.1} & 2.66\autodeltatwoHB{+0.12} \\
        & ClearSight & 57.3\autodeltaoneLB{-2.1} & 2.52\autodeltatwoHB{-0.02} \\
        & VTI & 51.0\autodeltaoneLB{-8.4} & 2.72\autodeltatwoHB{+0.18} \\
        & Nullu & 54.2\autodeltaoneLB{-5.2} & 2.70\autodeltatwoHB{+0.16} \\
        & \cellcolor{gray!15}DeP(Ours) & \cellcolor{gray!15}\textbf{49.0}\autodeltaoneLB{-10.4} & \cellcolor{gray!15}\textbf{2.83}\autodeltatwoHB{+0.29} \\
        \midrule
        \multirow{7}{*}{\textbf{InstructBLIP}}
        & \cellcolor{gray!15}Origin & \cellcolor{gray!15}93.8\autodeltaoneLB{0.0} & \cellcolor{gray!15}0.56\autodeltatwoHB{0.00} \\
        & VCD & 90.6\autodeltaoneLB{-3.2} & 0.61\autodeltatwoHB{+0.05} \\
        & CICD & 87.5\autodeltaoneLB{-6.3} & 0.74\autodeltatwoHB{+0.18} \\
        & ClearSight & 88.5\autodeltaoneLB{-5.3} & 0.70\autodeltatwoHB{+0.14} \\
        & VTI & 91.7\autodeltaoneLB{-2.1} & 0.55\autodeltatwoHB{-0.01} \\
        & Nullu & 85.4\autodeltaoneLB{-8.4} & 0.92\autodeltatwoHB{+0.36} \\
        & \cellcolor{gray!15}DeP(Ours) & \cellcolor{gray!15}\textbf{83.3}\autodeltaoneLB{-10.5} & \cellcolor{gray!15}\textbf{1.14}\autodeltatwoHB{+0.58} \\
        \bottomrule
        \end{tabular}
        \label{tab:mmhal_main}
    \end{subtable}
    \hfill
    \begin{subtable}[t]{0.35\textwidth}
        \centering
        \caption{Detailed scores on categories}
        \includegraphics[width=\linewidth]{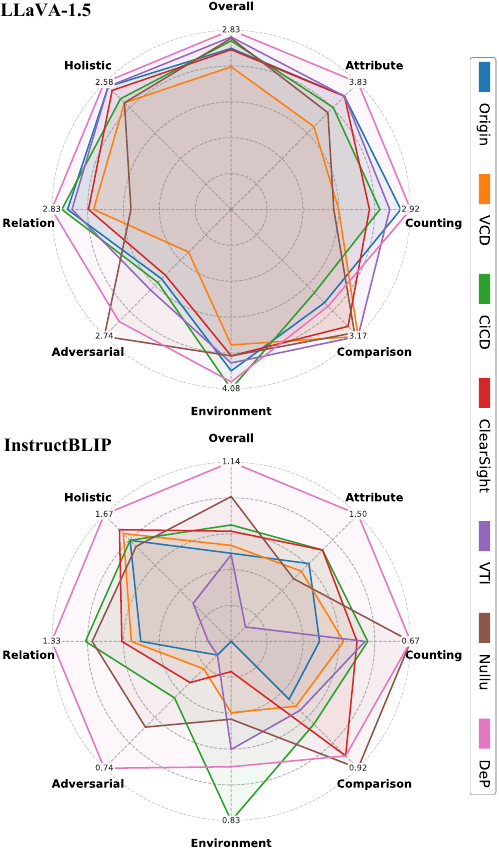}
        \label{fig:mmhal_radar}
    \end{subtable}
\end{table}

\subsubsection{Performance on MMHal-Bench}
Tab.~\ref{tab:mmhal_main} demonstrates that DeP effectively reduces the hallucination rate and improves the generation quality score on the MMHal-Bench dataset, achieving state-of-the-art performance. For LLaVA-1.5, DeP substantially decreases the hallucination rate from \textbf{59.4} to \textbf{49.0} while increasing the comprehensive score from \textbf{2.54} to \textbf{2.83}.
Notably, the baseline model InstructBLIP faces immense challenges on this dataset, exhibiting an original hallucination rate as high as \textbf{93.8} and a score of only \textbf{0.56}. Nevertheless, DeP still exhibits excellent intervention efficacy, successfully reducing the hallucination rate of InstructBLIP to \textbf{83.3} and raising the score to \textbf{1.14}.

As illustrated in the Tab.~\ref{fig:mmhal_radar}, we present a multidimensional performance comparison of various methods across different fine-grained dimensions of MMHal-Bench. Whether on LLaVA-1.5 or InstructBLIP, DeP demonstrates the best performance across most evaluation dimensions. Particularly on InstructBLIP, the performance polygon of the original model almost shrinks to the center. After incorporating the DeP framework, its performance boundaries are substantially expanded outward in all directions. This indicates that DeP effectively guides the model to capture and remain faithful to local visual details of the image while mitigating the interference of text co-occurrence.

\subsection{Ablation Studies}

\begin{table}[t]
\centering
\caption{Ablation study of DeP components. \textbf{(a)} Results of core components on POPE and MMHal benchmarks. \textbf{(b)} Effectiveness of our $N_{\mathrm{eff}}$ strategy against random perturbation ratios $p$ on MMHal. Delta denotes the improvement compared to Origin.}
\label{tab:ablation_combined}

% ----------------- 子表 (a) -----------------
\begin{subtable}[t]{0.60\linewidth}
\centering
\caption{Ablation study on POPE and MMHal benchmarks}
\footnotesize
\setlength{\tabcolsep}{1pt}
\vspace*{3pt}
\begin{tabular}{lcccc}
\toprule
\multirow{2}{*}{Method} & \multicolumn{2}{c}{POPE (Popular)} & \multicolumn{2}{c}{MMHal} \\
\cmidrule(lr){2-3} \cmidrule(lr){4-5}
 & Acc$\uparrow$ & F1$\uparrow$ & Halluc. Rate$\downarrow$ & Score$\uparrow$ \\
\midrule
\rowcolor{gray!15}Origin & 77.0\autodelta{0.0} & 79.7\autodelta{0.0} & 59.4\autodeltaoneLB{0.0} & 2.54\autodeltatwoHB{0.0} \\
w/o PC
 & 78.6\autodelta{1.6}
 & 80.9\autodelta{1.2}
 & 55.2\autodeltaoneLB{-4.2}
 & 2.65\autodeltatwoHB{0.11} \\
w/o VE
 & 80.2\autodelta{3.2}
 & 81.6\autodelta{1.9}
 & 53.1\autodeltaoneLB{-6.3}
 & 2.72\autodeltatwoHB{0.18} \\
\rowcolor{gray!15} DeP(Ours)
 & \textbf{82.8}\autodelta{5.8}
 & \textbf{84.7}\autodelta{5.0}
 & \textbf{49.0}\autodeltaoneLB{-10.4}
 & \textbf{2.83}\autodeltatwoHB{0.29} \\
\bottomrule
\end{tabular}
\label{tab:module_ablation_a}
\end{subtable}%
\hfill
% ----------------- 子表 (b) -----------------
\begin{subtable}[t]{0.38\linewidth}
\centering
\caption{Comparison of the $N_{\mathrm{eff}}$ strategy}
\footnotesize
\setlength{\tabcolsep}{1pt}
\vspace*{-4pt}
\begin{tabular}{lcc}
\toprule
$p$ & Halluc. Rate$\downarrow$ & Score$\uparrow$ \\
\midrule
% Origin & 59.4\autodeltaoneLB{0.0} & 2.54\autodeltatwoHB{0.0} \\
% \midrule
0.00 & 53.1\autodeltaoneLB{-6.3} & 2.60\autodeltatwoHB{0.06} \\
0.25 & 51.0\autodeltaoneLB{-8.4} & 2.64\autodeltatwoHB{0.10} \\
0.50 & 52.1\autodeltaoneLB{-7.3} & 2.61\autodeltatwoHB{0.07} \\
0.75 & 50.0\autodeltaoneLB{-9.4} & 2.74\autodeltatwoHB{0.20} \\
1.00 & 51.0\autodeltaoneLB{-8.4} & 2.74\autodeltatwoHB{0.20} \\
\midrule
$N_{\mathrm{eff}}$ & \textbf{49.0}\autodeltaoneLB{-10.4} & \textbf{2.83}\autodeltatwoHB{0.29} \\
\bottomrule
\end{tabular}
\label{tab:module_ablation_b}
\end{subtable}

\end{table}

% \begin{table}[t]
% \centering
% \caption{Ablation study of DeP components on POPE and MMHal benchmarks. Delta denotes the improvement compared to Origin.}
% \setlength{\tabcolsep}{5pt}
% \begin{tabular}{lcccc}
% \toprule
% \multirow{2}{*}{Method} & \multicolumn{2}{c}{POPE (Popular)} & \multicolumn{2}{c}{MMHal} \\
% \cmidrule(lr){2-3} \cmidrule(lr){4-5}
%  & Acc & F1 & Halluc. Rate & Score \\
% \midrule
% Origin & 77.0\autodelta{0.0} & 79.7\autodelta{0.0} & 59.4\autodeltaoneLB{0.0} & 2.54\autodeltatwoHB{0.0} \\
% w/o Prior Correction
%   & 78.6\autodelta{1.6}
%   & 80.9\autodelta{1.2}
%   & 55.2\autodeltaoneLB{-4.2}
%   & 2.65\autodeltatwoHB{0.11} \\
% w/o Visual Enhancement
%   & 80.2\autodelta{3.2}
%   & 81.6\autodelta{1.9}
%   & 53.1\autodeltaoneLB{-6.3}
%   & 2.72\autodeltatwoHB{0.18} \\
% DeP(Ours)
%   & 82.8\autodelta{5.8}
%   & 84.7\autodelta{5.0}
%   & 49.0\autodeltaoneLB{-10.4}
%   & 2.87\autodeltatwoHB{0.33} \\
% \bottomrule
% \end{tabular}
% \label{tab:module_ablation}
% \end{table}

\begin{figure}[t]
    \centering
    \begin{subfigure}[t]{0.48\textwidth}
        \centering
        \includegraphics[width=\linewidth]{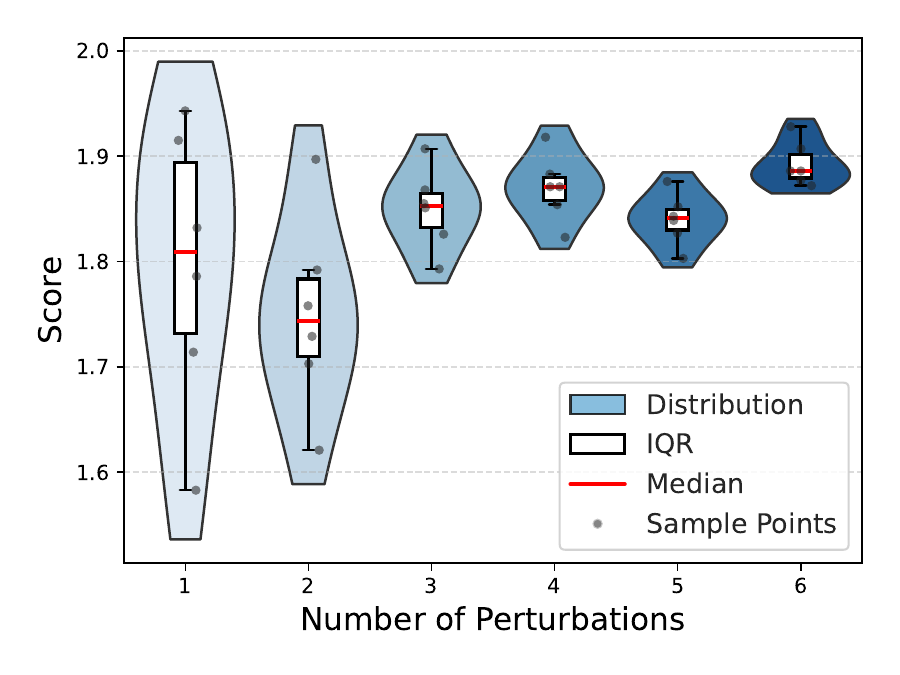}
        \caption{Overall Score vs. Perturbation Count}
        \label{fig:perturbation_box}
    \end{subfigure}\hfill
    \begin{subfigure}[t]{0.48\textwidth}
        \centering
        \includegraphics[width=\linewidth]{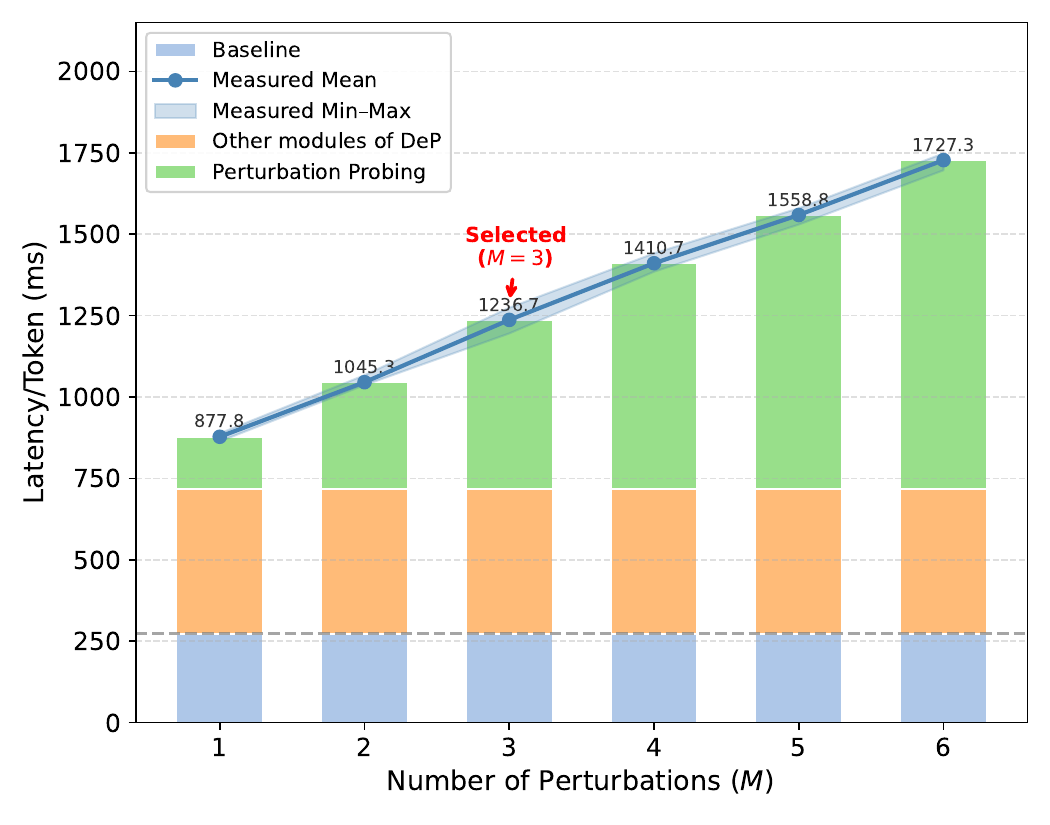}
        \caption{Inference Latency vs. Perturbation Count}
        \label{fig:perturbation_latency}
    \end{subfigure}

    \caption{Sensitivity analysis of perturbation count $M$. \textbf{(a)} Score variance decreases as $M$ increases, indicating improved stability in visual decoupling. \textbf{(b)} Inference latency scales linearly with $M$.}
    \label{fig:perturbation_analyse}
\end{figure}

\subsubsection{Module Ablation}

To deeply investigate the contributions of each core component in the DeP framework, we conduct detailed ablation studies on the POPE (Popular) and MMHal benchmarks using LLaVA-1.5 as the baseline. The experimental results are presented in the Tab.~\ref{tab:module_ablation_a}.

When removing the language prior correction module (\textbf{w/o PC}), the F1 score of the model on POPE increases from the baseline of \textbf{79.7} to \textbf{80.9}, and the hallucination rate on MMHal-Bench decreases from \textbf{59.4} to \textbf{55.2}. This indicates that our designed visual evidence decoupling and enhancement module can partially compel the model decisions to return to visual facts, preliminarily alleviating the weight imbalance between vision and language.
On the other hand, removing the visual enhancement module (\textbf{w/o VE}) results in a more substantial performance improvement. The POPE F1 score reaches \textbf{81.6}, and the MMHal hallucination rate drops significantly to \textbf{53.1}. This demonstrates that counteracting the probability skew caused by text co-occurrence in the decoding output space serves as an extremely efficient strategy to tackle high-intensity language prior interference.

\paragraph{Perturbation Mode Selection.}
We replace our adaptive $N_{\mathrm{eff}}$ selection with a random baseline, applying the aggressive $\mathcal{T}_3$  with a fixed probability $p$. Tab.~\ref{tab:module_ablation_b} shows that while static probabilities alleviate hallucinations compared to the Origin, they yield sub-optimal and fluctuating results (e.g., $p=1$ underperforms $p=0.75$). In contrast, our adaptive $N_{\mathrm{eff}}$ strategy outperforms all static settings by precisely applying intense probing only during high model uncertainty, preserving semantic stability elsewhere.
Furthermore, we find that DeP yields consistently strong performance when adjusting the $N_{\mathrm{eff}}$ threshold $\delta$ within a certain range, indicating good tolerance to this hyperparameter. 
A detailed analysis of the $N_{\mathrm{eff}}$ strategy is provided in the supplementary materials.

\begin{figure}[t]
    \centering
    \begin{subfigure}[t]{0.48\textwidth}
        \centering
        \includegraphics[width=\linewidth]{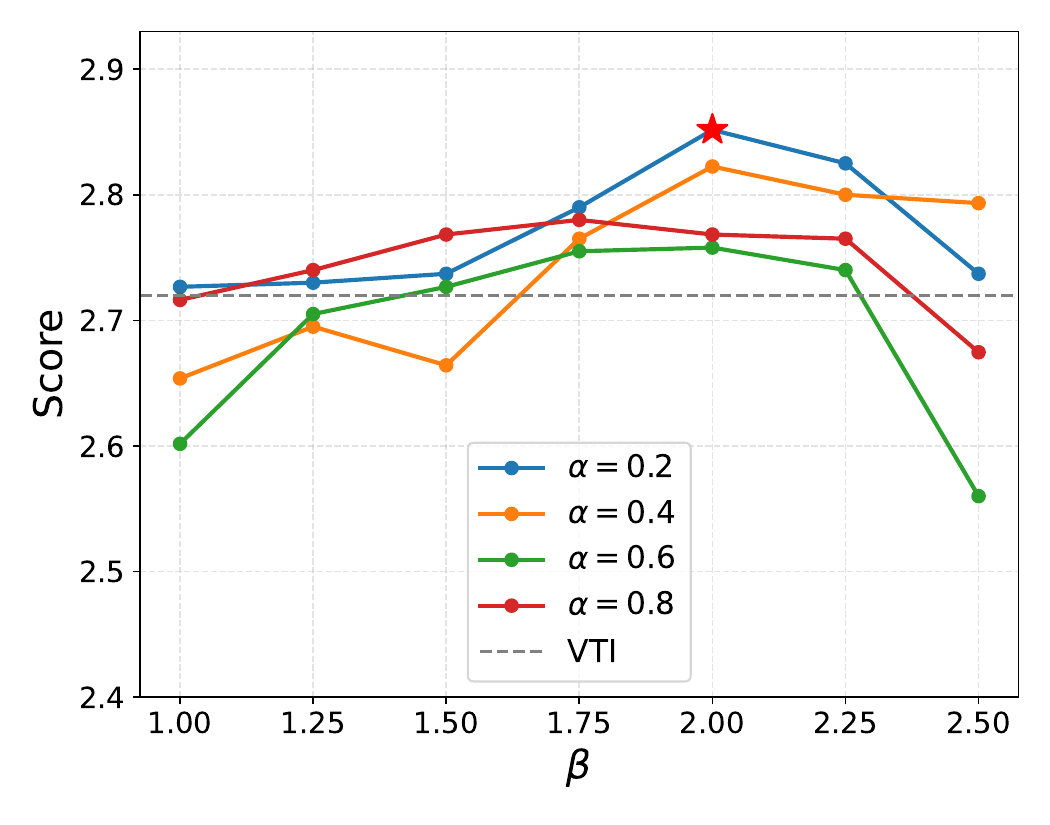}
        \caption{Overall Score vs. $\beta$ under different $\alpha$}
        \label{fig:alpha_gamma_score}
    \end{subfigure}\hfill
    \begin{subfigure}[t]{0.48\textwidth}
        \centering
        \includegraphics[width=\linewidth]{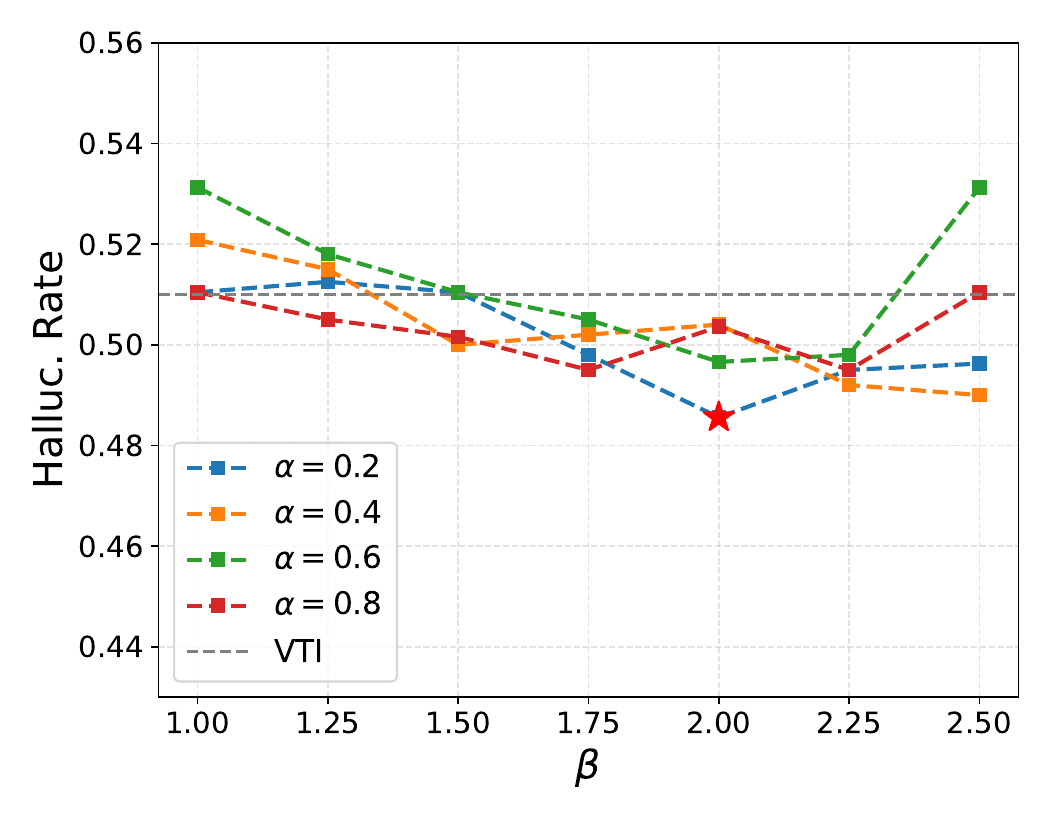}
        \caption{Hallucination Rate vs. $\beta$ under different $\alpha$}
        \label{fig:alpha_gamma_hallrate}
    \end{subfigure}

    \caption{Sensitivity of DeP performance to $\beta$ across different $\alpha$ values with $\gamma$ fixed to $0.3$. We report overall score \textbf{(a)} and hallucination rate \textbf{(b)}. The best-performing configuration is marked with \textbf{red star}. The dashed line indicates the results of the current best method (VTI).}
    \label{fig:hyperparam_two}
\end{figure}

\subsubsection{Hyperparameter Analysis}

We conduct all analyses by integrating DeP into LLaVA-1.5 and evaluating on MMHal-Bench. To reduce randomness and ensure stable conclusions, we repeat each configuration over multiple independent runs and report the mean performance.

\paragraph{Probe Sensitivity.}
As illustrated in Fig.~\ref{fig:perturbation_box}, smaller values ($M \le 2$) fail to form a reliable multi-sample distribution, resulting in high score variance and instability. Increasing $M$ significantly compresses this variance, yielding more consistent mitigation of language priors. However, Fig.~\ref{fig:perturbation_latency} demonstrates that inference latency scales strictly linearly with each additional perturbation. We observe that performance stability plateaus at $M=3$. Further increasing $M$ provides marginal stability gains but incurs prohibitive computational costs. Thus, we set $M=3$ as the optimal sweet spot to balance robust visual decoupling with acceptable decoding speed.

\paragraph{Intervention Intensity.} 
To evaluate the robustness of our dual-space correction, we investigate the hyperparameters governing spatial visual calibration ($\alpha, \gamma$) and semantic output guidance ($\beta$). Since preliminary experiments show the model is relatively insensitive to $\gamma$, we experimentally fix $\gamma = 0.3$ (see supplementary materials for detailed results). We further analyze the interplay between the visual evidence enhancement factor $\alpha \in \{0.2, 0.4, 0.6, 0.8\}$ and the language prior penalty $\beta \in \{1.00, 1.25, 1.50, 1.75, 2.00, 2.25, 2.50\}$. As illustrated in Fig.~\ref{fig:hyperparam_two}, a conservative visual enhancement of $\alpha = 0.2$ consistently yields the good performance. Aggressively increasing $\alpha$ (e.g., to $0.6$ or $0.8$) results in high variance and degraded scores, suggesting that over-amplifying visual features can distort the latent semantic representations of the LLM. Concurrently, increasing $\beta$ effectively suppresses hallucinations, culminating in a global optimum at $\beta = 2.0$. However, pushing the penalty too far ($\beta > 2.0$) disrupts natural linguistic transitions and causes over-correction, leading to a performance decline.

\section{Conclusion}

In this paper, we specifically tackle a critical bottleneck in Multimodal Large Language Models (MLLMs): inference-time hallucinations induced by language priors. We reveal that such prior-driven hallucinations fundamentally manifest as the hypersensitivity of visual grounding to surface-level textual phrasing during decoding. To bound and address this specific problem, we propose \textbf{Decoding by Perturbation (DeP)}, a training-free framework that isolates and mitigates language prior interference through controlled textual interventions.

DeP operates through a closed loop of detection and correction to reduce hallucinations. It employs a dynamic probe with multi-level textual perturbations to elicit latent language priors. By statistically analyzing cross-modal attention variance across these perturbations, DeP decouples stable visual evidence from textually drifting suspicious regions. This allows us to calibrate features in the hidden state space. Furthermore, to handle linguistic biases that cannot be mapped to specific regions, DeP formulates an interpretable prior drift direction using logits statistics to counteract probability skews stemming from textual co-occurrences.

\paragraph{Limitations.}
While the DeP framework effectively mitigates language prior-driven hallucinations, it exhibits certain limitations. First, the efficacy of our dynamic text perturbation probe is affected by the specificity of the input prompt. For highly generic or open-ended queries, such as ``\textit{Describe the image in detail},'' the prompt lacks distinct semantic anchors. In such scenarios, applying multi-level textual perturbations struggles to elicit meaningful attention variance without fundamentally altering the task instruction, thereby limiting the framework's ability. Second, although we implement a visual mutual exclusion constraint to preserve semantic integrity during probing, the act of textual perturbation inevitably introduces new, unforeseen conditional biases into the hidden states rather than strictly isolating existing language priors. Future research could explore more fine-grained, adaptive perturbation strategies or continuous latent-space probes to mitigate these side effects.

% \clearpage  % TODO FINAL: This \clearpage needs to be removed from both review and camera-ready versions.

% \section*{Acknowledgements}
% Please insert your acknowledgments here.

\bibliographystyle{splncs04}
\bibliography{main}
\end{document}